\documentclass[conference]{IEEEtran}
\IEEEoverridecommandlockouts    

\usepackage{multirow}
\usepackage{lipsum}
\usepackage{amsmath}
\usepackage{amssymb}
\usepackage{multirow}
\usepackage[ruled,vlined]{algorithm2e}
\usepackage{graphicx}
\usepackage[table]{xcolor}
\graphicspath{{./Figures/}}
\usepackage[hyphens]{url}
\usepackage[hidelinks]{hyperref}
\usepackage[capitalise]{cleveref}

\usepackage{tikz}

\newcommand\copyrighttext{%
  \centering\footnotesize 
  This work has been submitted to IEEE for possible publication. Copyright may be transferred without notice, after which this version may no longer be accessible.
  }
\newcommand\copyrightnotice{%
\begin{tikzpicture}[remember picture,overlay]
\node[anchor=south,yshift=-2cm] at (current page.north) {\fbox{\parbox{\dimexpr0.9\textwidth-\fboxsep-\fboxrule\relax}{\copyrighttext}}};
\end{tikzpicture}%
}

\title{\LARGE \bf Will it stop? Comparing AEBS and ACC Effectiveness\\ in Commercial Passenger Vehicles}

\author{
	\parbox{\textwidth}{%
		\centering
		Marcello Cellina$^{1}$, Riccardo Don\`a$^{1}$, Stefano Favelli$^{1}$, Fabrizio Re$^{1}$, Davide Maggi$^{1}$,\\ Jessica Leoni$^{1}$, Espedito Rusciano$^{2}$, Akos Kriston$^{1}$%
	}%
	\thanks{$^{1}$Joint Research Centre for the European Commission, Ispra, IT
		{\tt\small name.surname@ec.europa.eu}}
	\thanks{$^{2}$Joint Research Centre for the European Commission, Petten, NL
		{\tt\small name.surname@ec.europa.eu}}%
}

\begin{document}
	
\maketitle
\copyrightnotice

\thispagestyle{empty}
\pagestyle{empty}

\begin{abstract} 
Advanced Driver Assistance Systems (ADAS) operating on longitudinal control are increasingly deployed in passenger vehicles.
However, their performance and safety impact outside standardized testing conditions remain insufficiently characterized.
This study investigates the behavior of commercial ADAS in Car-to-Car Rear Stationary (CCRs) scenarios, comparing it with Advanced Emergency Braking Systems (AEBS) and Adaptive Cruise Control (ACC). In particular, we evaluate the responses of these systems on two passenger vehicles through a proving-ground campaign combining regulation scenarios and controlled variations. Experimental results reveal different longitudinal intervention strategies: AEBS exhibits a predominantly reactive braking response concentrated at shorter time-to-collision values, whereas ACC provides a more progressive longitudinal response over a wider range of conditions. For AEBS, test outcomes were further analyzed to investigate factors associated with unsuccessful interventions: initial speed alone proved not statistically significant, in contrast to normalized speed variability during the target approach phase, suggesting that driver behavior may impact AEBS performance. However, additional testing under controlled conditions remains necessary to consolidate such findings. The results demonstrate the additional value of controlled proving-ground experiments for characterizing commercial ADAS behavior beyond conventional binary pass/fail assessment. The testing data is accessible at \url{https://data.jrc.ec.europa.eu/dataset/406183af-97f9-471e-b86c-07e6c68e2ed8}.

\end{abstract}

\section{Introduction}
\label{sec:introduction}
Passenger cars are increasingly equipped with Advanced Driver Assistance Systems (ADAS), including safety-oriented functions such as Advanced Emergency Braking Systems (AEBS) and comfort-oriented functions such as Adaptive Cruise Control (ACC). Although both of these functions influence the longitudinal dynamics of the vehicle, they serve different purposes. However, their operating envelopes are progressively converging: the increasing activation speeds of AEBS and the wider operational ranges of modern ACC-assisted driving systems result in a growing domain overlap, in which longitudinal interventions from either system may occur. Therefore, characterizing how these systems behave within and beyond standardized operating conditions is necessary for assessing their effectiveness and potential limitations.

Independent evaluation of commercial ADAS supports evidence-based safety assessment and policymaking \cite{dona2024experimental}. Consumer organizations (e.g., Euro NCAP and IIHS) and research institutions conduct such assessments \cite{euroncap2024protocolAEB,skokan2025field,Zou2025152,schwall2014testing}, providing manufacturer-independent evidence of system performance and safety implications \cite{fildes2015effectiveness,kullgren2023effect}.

However, experimental campaigns are constrained by vehicle availability, proving-ground access, instrumentation, and data-processing effort. Moreover, standardized protocols prioritize repeatability and comparability, potentially leaving ADAS behavior under non-standard conditions insufficiently characterized.

This paper presents an experimental characterization of the ADAS behavior of two recent passenger vehicles available on the market. We conducted a proving-ground testing campaign for Car-to-Car Rear Stationary (CCRs) scenarios, including both regulatory test conditions and controlled variations. 
The tests covered both ACC, with longitudinal control managed by the vehicle, and AEBS, with the driver maintaining control until an emergency braking system intervened.
Two independent data acquisition systems recorded vehicle trajectories, enabling the analysis independently from the vehicles' internal signals. Rather than considering AEBS and ACC as isolated functions, this study investigates the longitudinal assistance suite as a whole, allowing their different intervention strategies to be compared under similar conditions.

For AEBS, we expand the analysis from a conventional pass/fail assessment to investigate the factors associated with unsuccessful interventions. In particular, we examine the relationship between the test outcome and selected kinematic and trajectory-related variables statistically, including approach speed and approach-speed variability.

\begin{figure}[t]
    \centering
    \includegraphics[width=.99\linewidth]{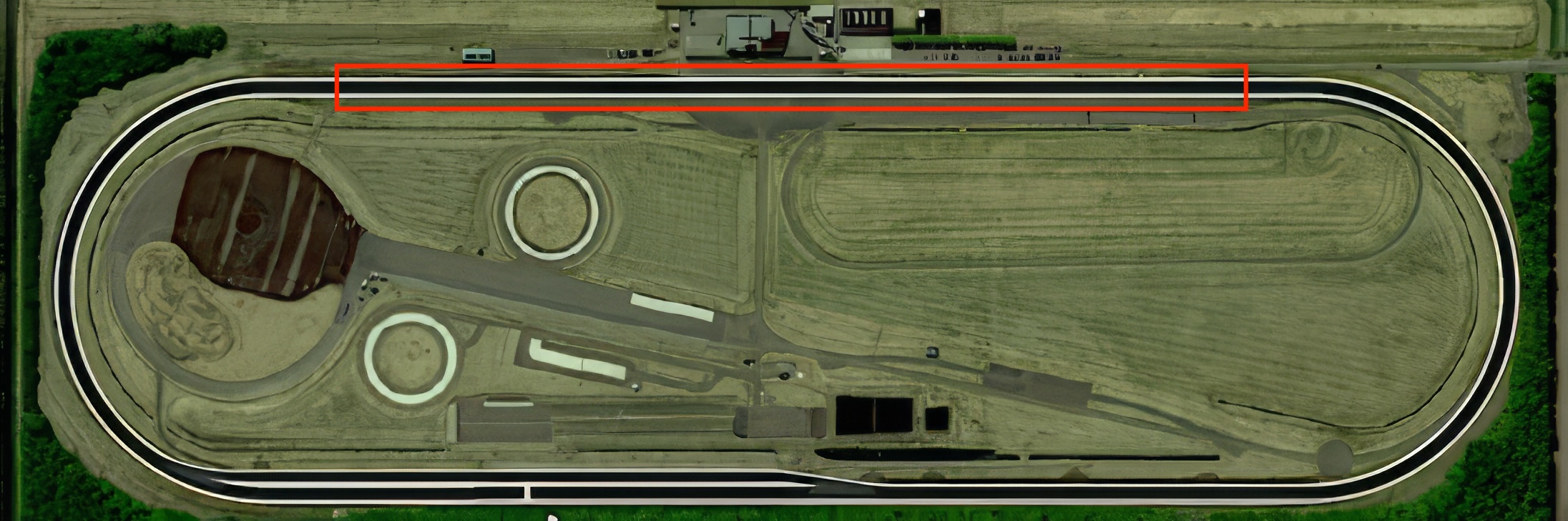}
    \caption{Overview of the RDW proving ground in Lelystad, the Netherlands,
    showing the test area used for the experimental campaign in red.}
    \label{fig:lelystad}
\end{figure}

The main contributions of this study are:
\begin{itemize}
    \item an empirical characterization of the longitudinal behavior of commercial ADAS across two passenger vehicles under standardized and systematically varied scenarios;
    \item a function-level comparison of the intervention characteristics and safety-related response of ACC and AEBS, aggregating results across the two vehicles;
    \item a statistical investigation of dynamical characteristics associated with unsuccessful AEBS interventions, going beyond a conventional binary pass/fail assessment.
\end{itemize}


\section{Methodology}\label{sec:methods}

\subsection{Testing Campaign}
The testing campaign was conducted in March 2026 at the Netherlands Vehicle Authority (RDW) proving ground in Lelystad, the Netherlands. \cref{fig:lelystad} shows the test track layout.

We performed two categories of rear-end scenarios along the straight section of the proving ground:
\begin{itemize}
    \item \textbf{CCRs:} the Vehicle Under Test (VUT) approaches a stationary dummy vehicle target at different speeds and lateral overlaps, with the considered ADAS function enabled.
    \item \textbf{Cut-out:} the VUT, with the considered ADAS function engaged, follows a lead vehicle that suddenly changes lane to reveal a stationary dummy target in the VUT's lane. Different longitudinal distances between the lead vehicle and the target, initial speeds, and ADAS configurations were investigated.
\end{itemize}

A test run was classified as successful when the relevant ADAS function was able to bring the VUT to a stop before reaching the target without intervention by the safety driver. To mitigate the risk of damage to the test vehicles and instrumentation in the event of a collision, a static Global Vehicle Target (GVT) was used as the surrogate target vehicle and the drivers were instructed to intervene and to avoid the impact. 

The test procedure was adapted to the characteristics of the ADAS function under investigation, corresponding to the two operating modes introduced in Section \ref{sec:introduction}:
\begin{itemize}
    \item \textbf{AEBS} (manual driving mode): the safety driver approached the target manually while maintaining the prescribed test speed and minimizing steering inputs, since steering activity may be interpreted by the driver monitoring system as a sign of driver engagement, leading to the suppression of the braking intervention. The driver performed an evasive steering maneuver only if AEBS had not intervened by the last safe point for evasive steering \cite{dona2023towards}. A cone positioned alongside the test path was used as a visual reference for the intervention point.
    \item \textbf{ACC} (assisted mode): the safety driver manually accelerated the VUT to the target speed and subsequently activated ACC or the ADAS system with the prescribed speed and time-headway settings. As in the AEBS tests, the driver performed an evasive maneuver only if the system had not initiated braking by the latest safe intervention point.
\end{itemize}

\cref{tab:TM_global} summarizes the distribution of test runs.

\begin{table}[htbp]
\renewcommand{\arraystretch}{1.2}
\caption{Distribution of test runs across scenarios, vehicles, and
investigated ADAS functions.}
\begin{center}
\begin{tabular}{|c|c|p{1.cm}|p{1.cm}|p{1.cm}|p{1.cm}|}
\hline
\multirow{3}{4em}{\textbf{Scenario}} &
\multirow{3}{4em}{\textbf{Vehicle}} &
\multicolumn{4}{c|}{\textbf{Test Runs}} \\
\cline{3-6}

& & \multicolumn{2}{c|}{\textbf{Nominal}} &
\multicolumn{2}{c|}{\textbf{Variations}} \\
\cline{3-6}

& & \textbf{\textit{AEBS}} &
\textbf{\textit{ACC}} &
\textbf{\textit{AEBS}} &
\textbf{\textit{ACC}} \\
\hline
\multirow[c]{3}{4em}{\centering\textbf{CCRs}} 
& Veh 1 & 7 & 15 & 6 & 15
\\
\cline{2-6} 
& Veh 2 & 16 & 11 & 11  & 15 \\
\cline{2-6} 
& \cellcolor{gray!20} \textit{Total} & \cellcolor{gray!20}  23 & \cellcolor{gray!20} 26 & \cellcolor{gray!20} 17 & \cellcolor{gray!20} 30 \\
\hline
\multicolumn{2}{|c|}{\cellcolor{gray!40} Speed range (km/h)} & \cellcolor{gray!40} 25 - 80 & \cellcolor{gray!40} 40 - 120 & \cellcolor{gray!40} - & \cellcolor{gray!40} - \\
\hline
\multirow[c]{3}{4em}{\centering\textbf{Cut out}}  
& Veh 1 & 3 & 6 & - & -\\
\cline{2-6} 
& Veh 2 & 6 & 12 & - & - \\
\cline{2-6} 
&  \cellcolor{gray!20}\textit{Total} & \cellcolor{gray!20} 9 & \cellcolor{gray!20} 18 & \cellcolor{gray!20} -  & \cellcolor{gray!20} -  \\
\hline
\multicolumn{2}{|c|}{\cellcolor{gray!40} Speed range (km/h)} & \cellcolor{gray!40} 40 - 80 & \cellcolor{gray!40} 40 - 80 & \cellcolor{gray!40} - & \cellcolor{gray!40} - \\
\hline
\end{tabular}
\label{tab:TM_global}
\end{center}
\end{table}

Although the campaign was conducted using two different vehicles, the analysis presented in this paper is carried out at the ADAS-function level. The results from the two vehicles are therefore aggregated, and the primary comparison is between AEBS and ACC. Accordingly, differences between individual vehicle implementations are beyond the scope of this study. ``Variations'' include different lateral offsets for the static target and night conditions, which are fully documented in the dataset. 

\subsection{Instrumentation \& Data Processing}\label{subsec:dataproc}
Two independent data acquisition systems were deployed during the testing campaign to provide measurement redundancy and enable cross-validation of the recorded vehicle trajectories and system responses:
\begin{itemize}
    \item \textbf{VBOX RTK:} a conventional proving-ground measurement system based on a high-accuracy GNSS receiver with RTK corrections, complemented by an inertial measurement unit (IMU).
    \item \textbf{IVEX Reference Sensor Kit:} a roof-mounted, LiDAR-camera-GNSS reference sensing system offering reduced setup time compared with conventional proving-ground instrumentation and no need for RTK setup.
\end{itemize}
The VBOX RTK measurements were therefore used as the reference for the analysis presented in this paper, as the LiDAR-based evaluation kit was later found to be less accurate than the RTK GNSS \cite{cellina2026beyond}.

Both systems recorded data at a sampling frequency of 100~Hz. For the cut-out scenarios, the measurements from the VUT and the lead vehicle were additionally synchronized to ensure consistent timestamp alignment between the two trajectories.

The VBOX RTK system provides high-accuracy trajectory measurements together with precomputed safety-related quantities, including Time-to-Collision (TTC) and bumper-to-bumper distance between the VUT and the target. For AEBS tests, the VBOX system was also used to record the acoustic warning associated with the system intervention.

\subsection{Safety Impact Estimation}
To characterize the safety implications of the observed ADAS responses, a set of kinematic and outcome-based metrics was considered. Since the investigated scenarios represent rear-end conflict situations, the TTC was used as a surrogate measure of collision risk:
\begin{equation}\label{eq:ttc}
    TTC = \frac{d}{v_{\textnormal{VUT}} - v_{\textnormal{target}}},
\end{equation}
where $d$ is the longitudinal distance between the VUT and the target and $v_{\textnormal{VUT}}$ and $v_{\textnormal{target}}$ are their respective longitudinal velocities. TTC was evaluated at two characteristic stages of the longitudinal response: at the onset of mild braking, corresponding to a deceleration of 2~m/s$^2$, and when the deceleration exceeded 5~m/s$^2$. These two points were selected to characterize the transition from the initial system response to more pronounced emergency braking and to facilitate comparison between the different longitudinal control strategies.

The success rate of AEBS and ACC was computed by aggregating the results across the two vehicles and grouping the tests according to their initial VUT speed. The speed ranges were selected to reflect the regulatory AEBS performance thresholds as defined in UN Regulation No.~152 \cite{UNR152}, which were adopted as a common reference for both functions. The selected four speed ranges, namely under 40~km/h, 40-60~km/h, 60-80~km/h, and above 80~km/h, are associated with the regulatory requirements, including the transition from collision avoidance (up to 40~km/h) to collision mitigation (up to 60~km/h).

To investigate the relationship between approach speed and AEBS outcome, we fitted a binary logistic regression model to the test-level results:
\begin{equation}\label{eq:logit_V}
    P(\textnormal{failure} \mid v) =
        \frac{1}
    {1 + e^{-(\alpha + \beta  v)}},
\end{equation}
where $v$ is the initial VUT speed and $\alpha$ and $\beta$ are the model coefficients. The logistic model provides a continuous estimate of the probability of an unsuccessful outcome as a function of approach speed. In this model, a failure was defined as a test in which the safety driver was required to intervene because the AEBS had not brought the VUT to a stop by the latest safe intervention point. 

To indicate the potential safety consequences associated with unsuccessful interventions, the estimated collision-speed distribution can additionally be related to injury risk using empirically derived relationships between relative collision speed $v_{\textnormal{striking}}$ and the probability of injury for different Maximum Abbreviated Injury Scale (MAIS) levels \cite{morden2026safe,hetz2026predicting}:
\begin{equation}\label{eq:mais}
    P(\textnormal{injury} \mid v_{\textnormal{striking}}) =
    \frac{1}
    {1 + e^{-(\alpha_\textnormal{MAIS} + \beta_\textnormal{MAIS} \, v_{\textnormal{striking}})}},
\end{equation}
where $\alpha_\textnormal{MAIS}$ and $\beta_\textnormal{MAIS}$ are parameters associated with the given MAIS level. Three classes of MAIS are considered: minor (MAIS 1), moderate (MAIS 2), and serious-or-fatal (MAIS 3+F) injuries. This relationship translates differences in residual collision speed into an indicative change in injury probability. In this study, the  $\alpha_\textnormal{MAIS}$ and $\beta_\textnormal{MAIS}$ coefficients corresponding to the ``unweighted'' approach in \cite{morden2026safe} were adopted.

Since the GVT was stationary in all tests, assuming that in the AEBS failure cases there would have been no mitigation by the driver, the estimated collision-speed variable was conservatively approximated by the initial VUT speed, i.e., $v_{\textnormal{striking}} \approx v$. The AEBS failure probability and the corresponding MAIS injury probability can then be combined to obtain an indicative counterfactual injury probability:

\begin{equation}\label{eq:injury_failure}
P_{\mathrm{injury}}(v) =
P(\mathrm{failure}\mid v) \cdot
P(\mathrm{injury}\mid v_{\textnormal{striking}} =v).
\end{equation}

This quantity represents the potential probability of the outcome of the injury considered if an unsuccessful AEBS intervention resulted in a collision at the initial approach speed, and is therefore interpreted as a counterfactual estimate of the impact of safety rather than an observed probability of injury.
    
\subsection{AEBS Failed Intervention Analysis}\label{subsec:fails}
The unsuccessful AEBS test runs were further analyzed to identify the dynamical characteristics that may have an association with the observed failures. In addition to the effect of approaching speed, investigated through the logistic regression in \cref{eq:logit_V}, we investigated the effect of the stability of the longitudinal approach. 

We define the Normalized Standard Deviation of the VUT speed $\sigma_{v,\mathrm{norm}}$, over the interval from 100~m to 20~m before the target, i.e., before any AEBS braking intervention, as in
\begin{equation}\label{eq:norm_speed_std}
    \sigma_{v,\mathrm{norm}} =
    \frac{\sigma_v}{\bar{v}},
\end{equation}
where $\sigma_v$ and $\bar{v}$ denote the standard deviation and mean VUT speed, respectively, computed over the selected approach interval.

The distributions of velocity and Normalized Standard Deviation were compared between successful and unsuccessful AEBS tests to understand which are the relevant features and which are not. Since the two groups differ in size and potentially in variance, statistical differences between the two groups were assessed using Welch's $t$-test, which does not require the assumption of equal variances between groups. 

\section{Results}\label{sec:results}
\subsection{System Behavior}\label{subsec:behavior}
\begin{figure}[t]
    \centering
    \includegraphics[width=.95\linewidth]{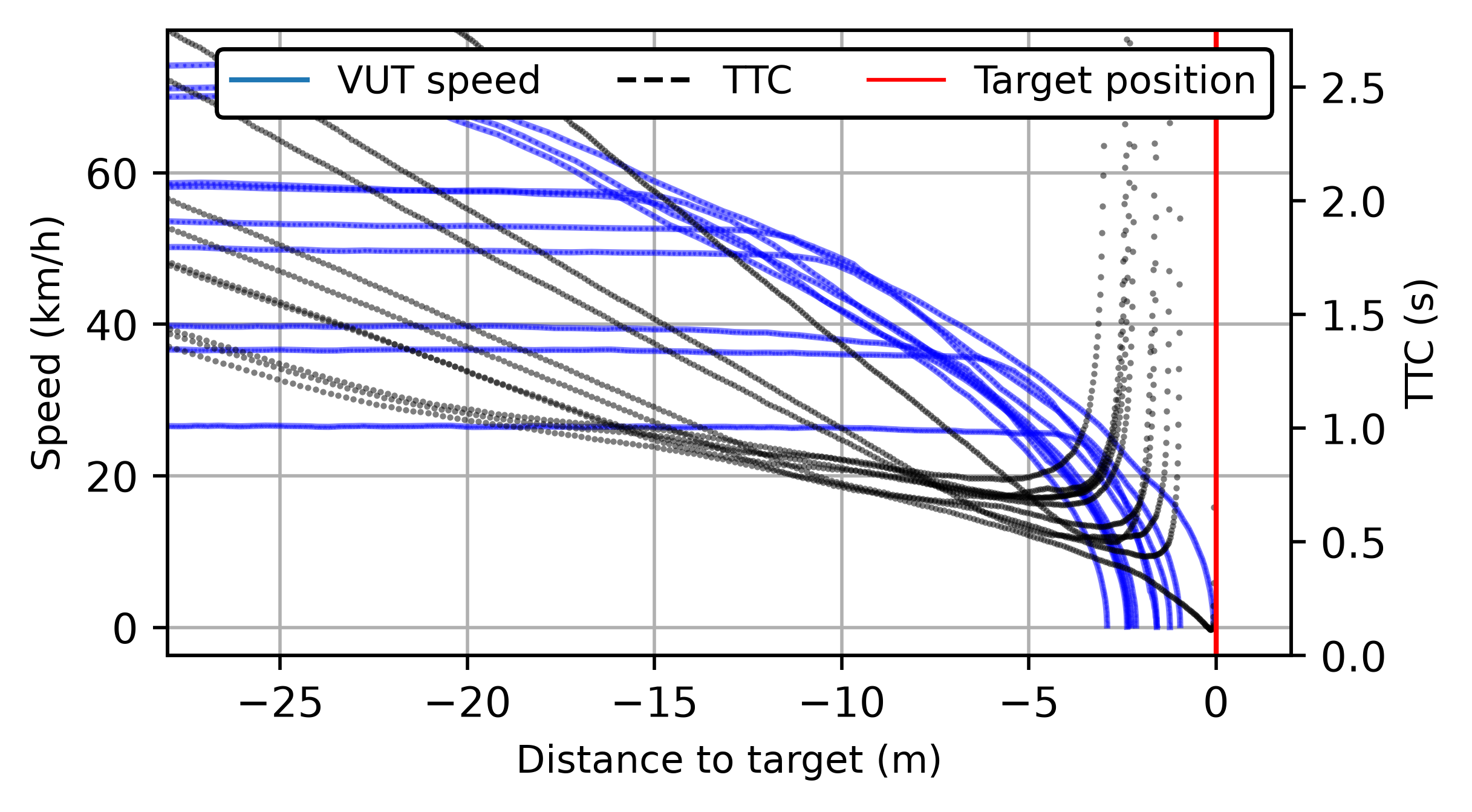}
    \caption{Representative successful AEBS speed and time-to-collision (TTC) profiles as a function of distance to the target. The red line indicates the target position.}
    \label{fig:speeds}
\end{figure}

\cref{fig:speeds} shows representative AEBS speed and TTC profiles as a function of the distance to the target. It can be noted that, despite differences in the initial approach conditions, the profiles exhibit a consistent late-stage intervention pattern, with the main reduction in vehicle speed occurring at short distances from the target. The corresponding TTC evolution illustrates the short time horizon over which the AEBS transitions from the initial response to maximum deceleration. These observations provide a first qualitative indication of the reactive nature of AEBS.

\begin{figure}[t]
    \centering
     \includegraphics[
        width=.95\linewidth,
        trim={0 20 0 0}, clip 
    ]{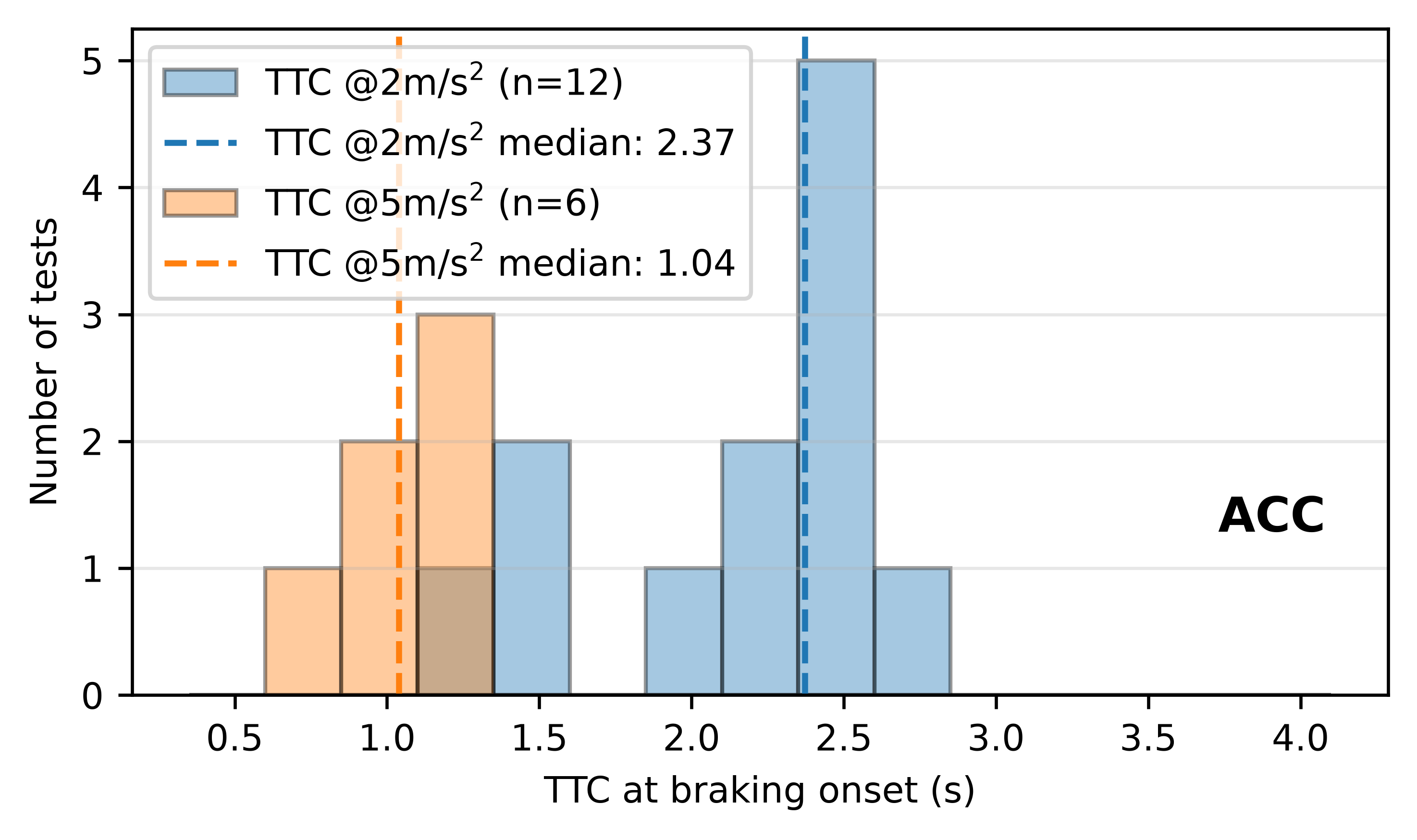}
    \quad
    \includegraphics[width=.95\linewidth]{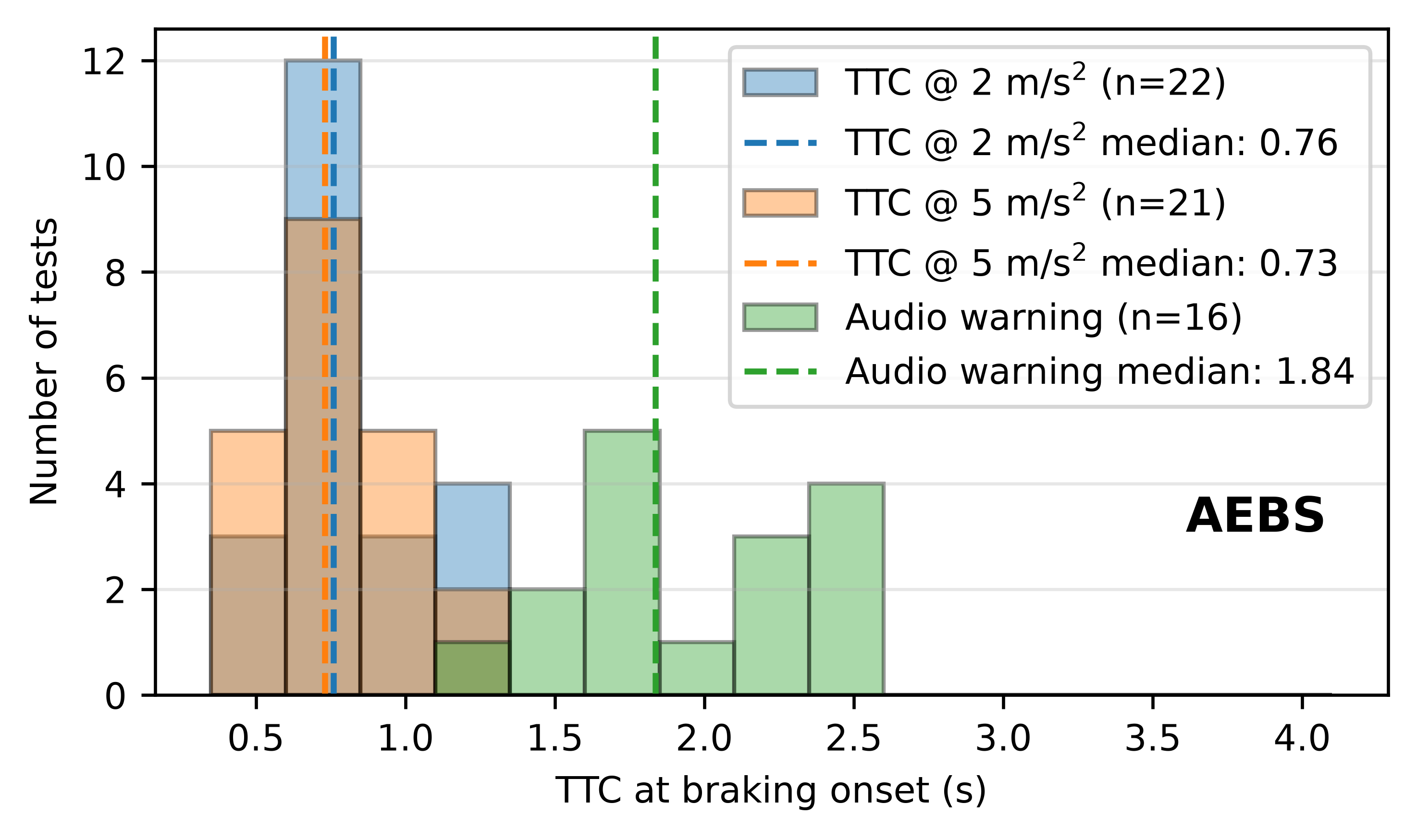}
    \caption{Distribution of time-to-collision (TTC) at the onset of mild
    ($\approx 2$~m/s$^2$) and pronounced ($>5$~m/s$^2$) braking for ACC (top) and AEBS (bottom). Dashed vertical lines indicate the corresponding median values. The AEBS panel additionally shows the 
    distribution of acoustic warning times.}
    \label{fig:hist}
\end{figure}

A more systematic comparison of the longitudinal response of AEBS and ACC is obtained from the TTC distributions at the onset of mild braking ($\approx 2$~m/s$^2$) and pronounced braking ($>5$~m/s$^2$), shown in \cref{fig:hist}. The corresponding acoustic-warning distribution is also reported for the AEBS tests.

The ACC results show a broad distribution of TTC values at the onset of mild braking, indicating substantial variability in when deceleration is initiated. A small number of cases exhibit an earlier response, with mild braking initiated at TTC values above 3.6~s. In contrast, when pronounced braking ($>5$~m/s$^2$) is required, the corresponding TTC values are concentrated in a considerably narrower range around 1~s. This indicates a progressive ACC response in which the initial longitudinal intervention can occur over a relatively broad range of traffic conditions, while stronger deceleration is concentrated closer to the conflict point.

The AEBS results exhibit a different intervention pattern. The TTC distributions associated with the 2~m/s$^2$ and $>5$~m/s$^2$ braking thresholds largely overlap and are concentrated at lower TTC values than those observed for ACC. This indicates that, once activated, AEBS transitions rapidly from its initial response to pronounced braking, consistent with its more reactive safety-oriented intervention strategy. The acoustic warning is triggered, with a median of approximately 1~s before the onset of braking, providing an earlier indication of the impending intervention.

\subsection{Success Probabilities \& Safety Impact}\label{subsec:res_success}
\begin{figure}[b!]
    \centering
    \includegraphics[
        width=.9\linewidth,
        trim={0 20 0 0}, clip 
    ]{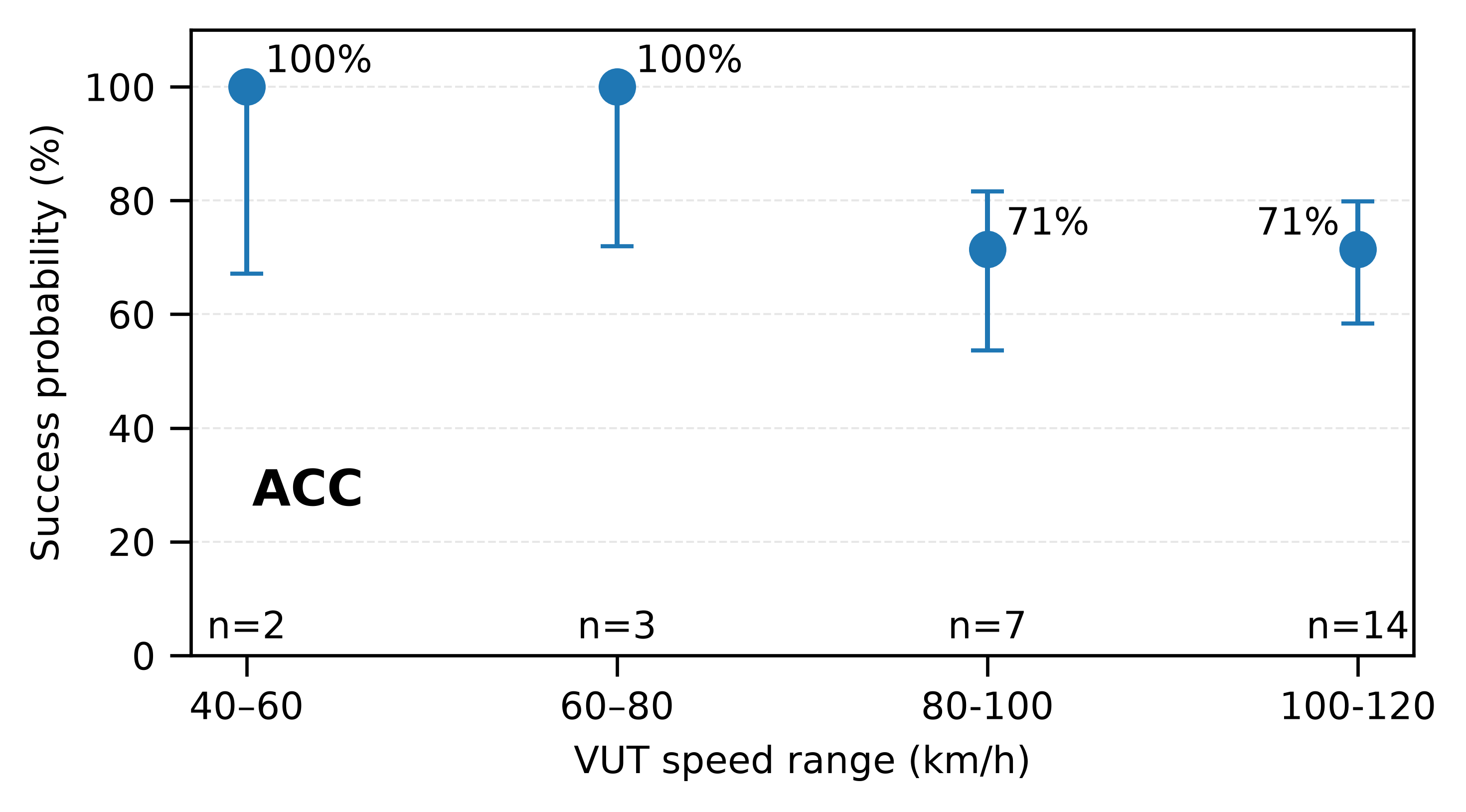}    
    \quad
    \includegraphics[width=.9\linewidth]{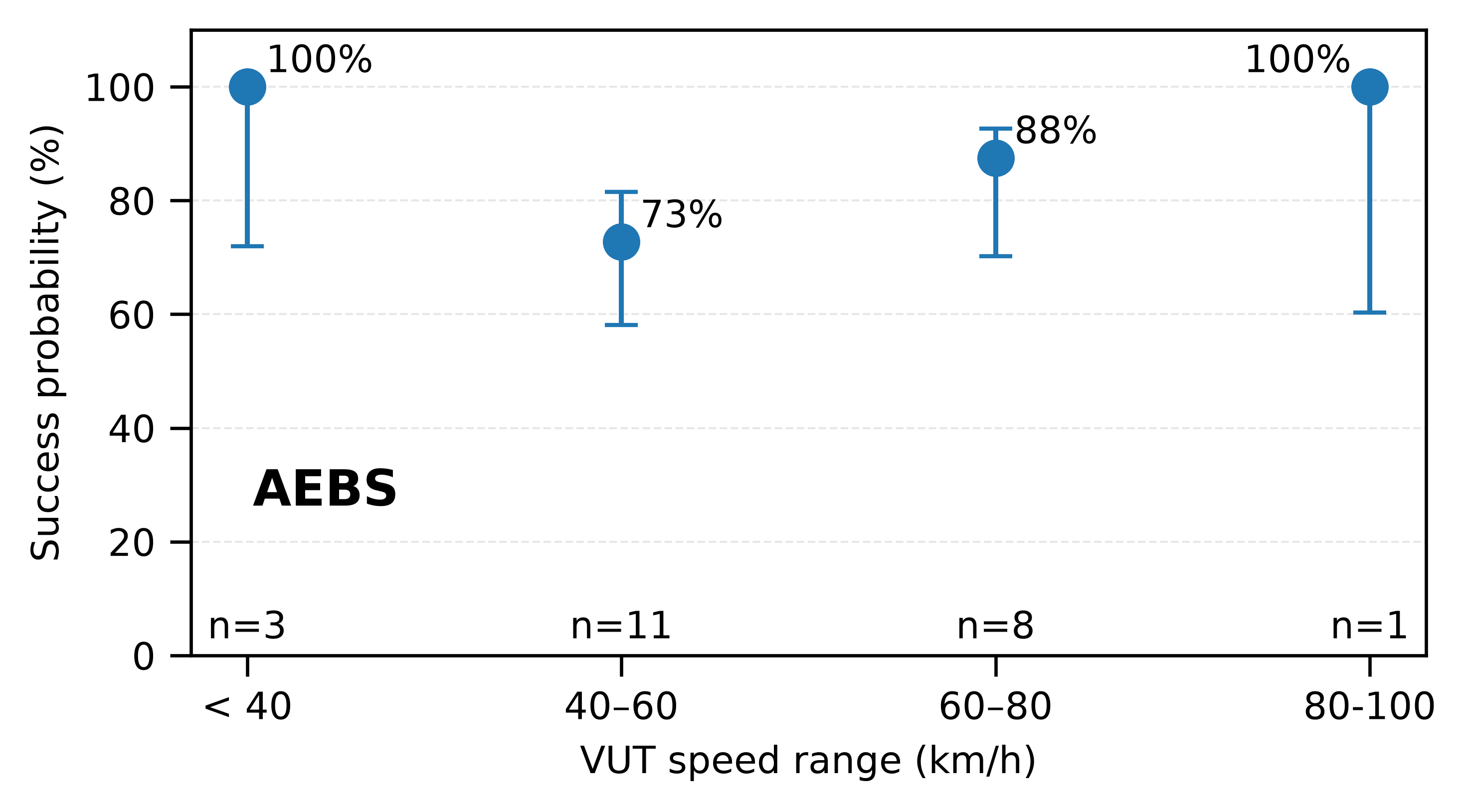}
    \caption{Observed system success proportions by initial VUT speed band for ACC (top) and AEBS (bottom). Error bars indicate the corresponding confidence intervals; $n$ denotes the number of test runs in each speed band.}
    \label{fig:success}
\end{figure}

The charts of the ADAS success rate as a function of the initial speed are shown in \cref{fig:success}, considering the ``nominal'' cases only. 

It shows that the ACC delivers almost 100\% success across the speed range considered, with some degradation of performance taking place starting from 80 km/h. The AEBS shows instead 100\% reliability in the sub-40~km/h area, in line with the provisions of UN Regulation No.~152.  
However, the performance deteriorates with speed, albeit in a couple of cases we reported successful intervention in the 60--80~km/h speed range, with one sample at 80 km/h being the highest speed we could confidently test the AEBS.

\begin{figure}[t]
    \centering
    \includegraphics[width=.9\linewidth]{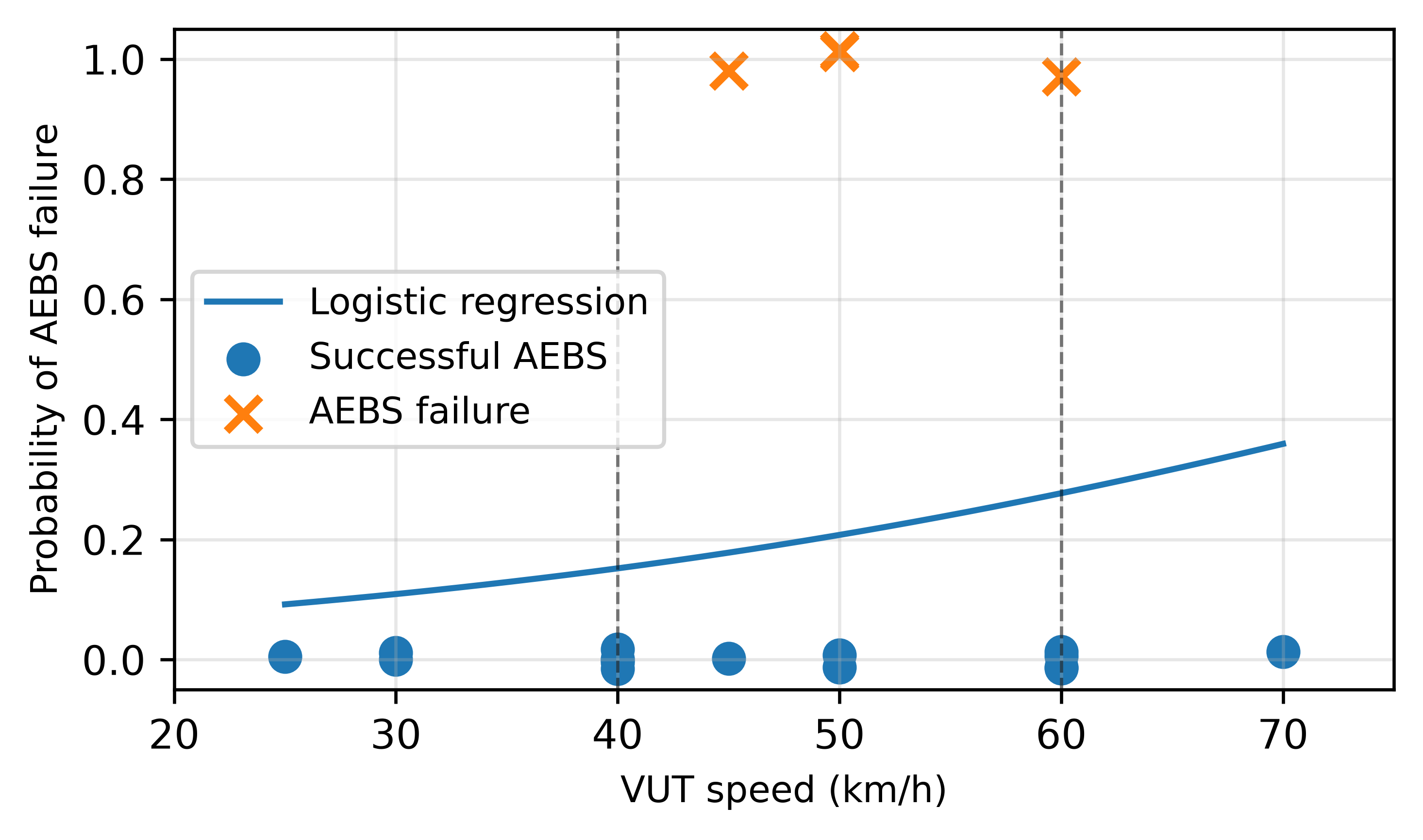}
    \caption{Test-level AEBS outcomes and fitted logistic regression model relating the probability of unsuccessful intervention to initial VUT speed.}
    \label{fig:logispeed}
\end{figure}

To further investigate the relationship between approach speed and AEBS outcome, a binary logistic regression was fitted using the test-level results of the 20 AEBS runs (analysis limited to initial speeds up to 70~km/h, since higher-speed conditions were tested with a single repetition only), of which 16 were successful and 4 resulted in a safety-driver intervention. The fitted model indicates a positive relationship between initial VUT speed and the probability of an unsuccessful intervention, as shown in \cref{fig:logispeed}. 
The estimated odds of AEBS failure increased by 3.9\% per 1 km/h increase in initial speed (odds ratio = 1.039, 95\% CI: 0.941--1.147); however, this association was not statistically significant ($p=0.450$). The model also provided limited explanatory power, with a McFadden pseudo-$R^2$ of 0.030, and the likelihood-ratio test likewise provided no evidence of a significant improvement over the null model ($p=0.439$).

The fitted model therefore provides insufficient evidence that initial approach speed alone significantly predicts AEBS failure under the tested conditions. Given the limited sample size, however, this should not be interpreted as evidence of no speed effect. 

\begin{figure}[t]
    \centering
    \includegraphics[
        width=.9\linewidth,
        trim={0 20 0 0}, clip 
    ]{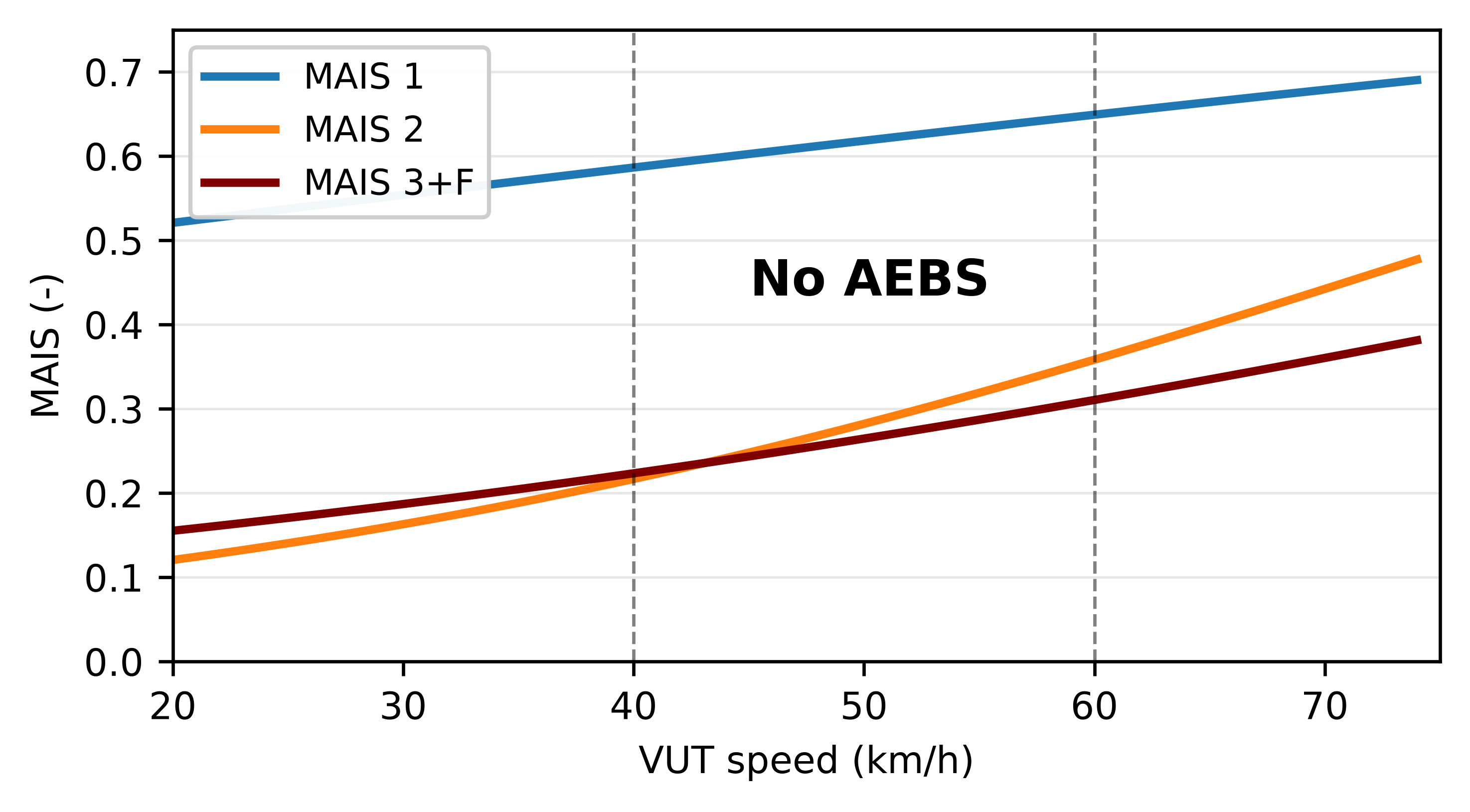}
    \quad
    \includegraphics[width=.9\linewidth]{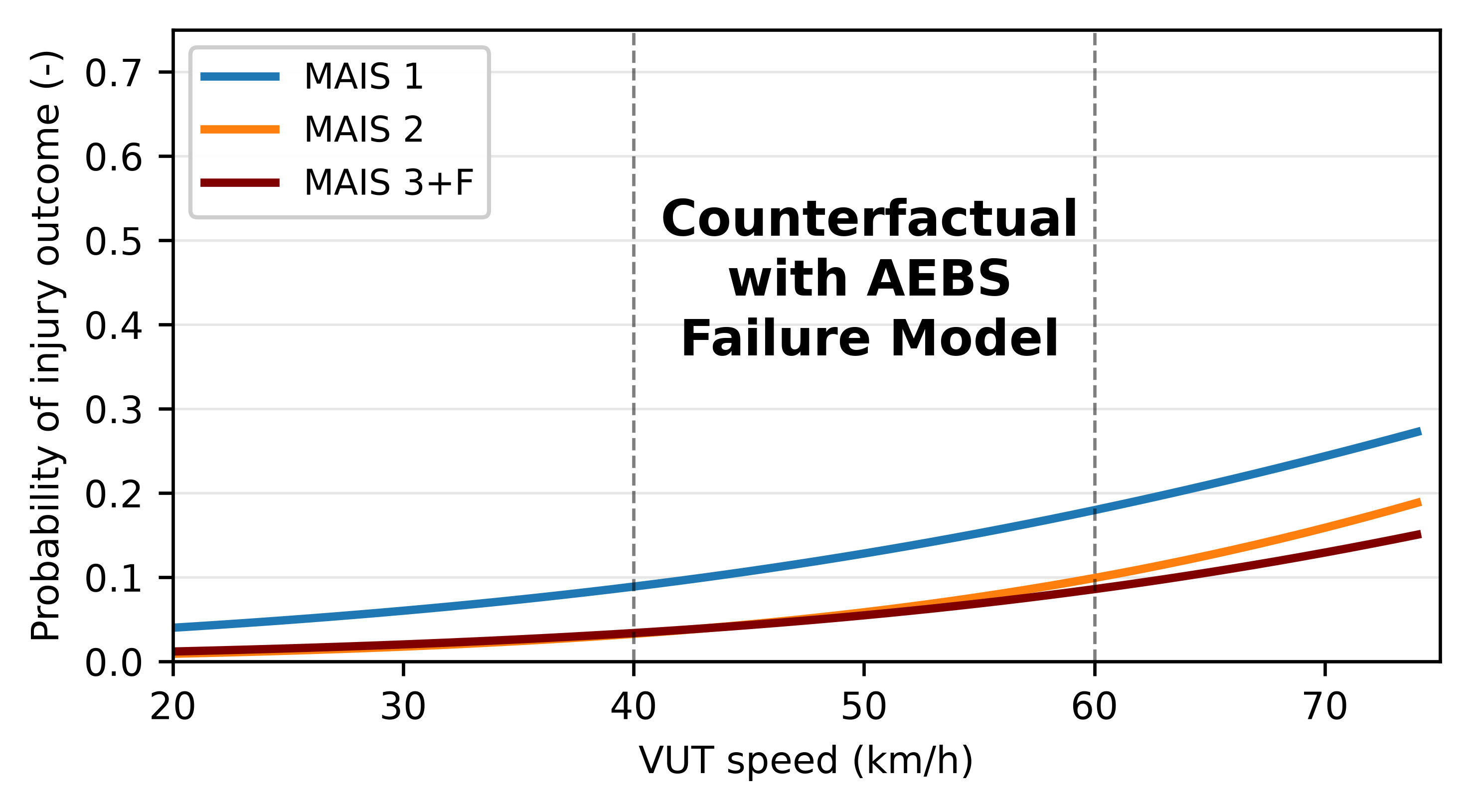}
    \caption{Raw MAIS estimated based on \cite{morden2026safe} study (top) and counterfactual injury probability accounting for AEBS failure (bottom).}
    \label{fig:mais}
\end{figure}

Despite the limited statistical power of the test-level analysis, the potential safety consequences of unsuccessful AEBS interventions were further investigated by relating the estimated failure probability to injury severity. To separate collision severity from AEBS performance, the injury-risk relationship of \cref{eq:mais} is first considered independently of the AEBS outcome. As shown in \cref{fig:mais} (top), the injury probability increases with collision speed for all MAIS levels.

The potential safety impact of an unsuccessful AEBS intervention can then be obtained by combining the conditional injury probability with the estimated probability of AEBS failure. The resulting counterfactual injury probability, given by \cref{eq:injury_failure}, is shown in \cref{fig:mais} (bottom). This quantity represents the probability of the considered injury outcome through the AEBS-failure pathway, and therefore accounts for both the likelihood of an unsuccessful intervention and the potential consequences of the resulting collision.

The calculation assumes that an unsuccessful intervention would result in a collision at the initial VUT approach speed, i.e., that no collision-speed mitigation occurs before impact. Since no physical collision occurs in the unsuccessful test runs, the corresponding collision severity cannot be directly measured. The resulting estimates should therefore be interpreted as counterfactual safety-impact estimates rather than observed or population-level injury risks. In particular, they do not account for the real-world exposure distribution of AEBS conflict situations and should not be interpreted as a direct estimate of the expected injury risk in real-world crashes.

\subsection{AEBS Failed intervention analysis}\label{subsec:res_fail}

\begin{figure}[b!]
    \centering
    \includegraphics[width=0.9\linewidth]{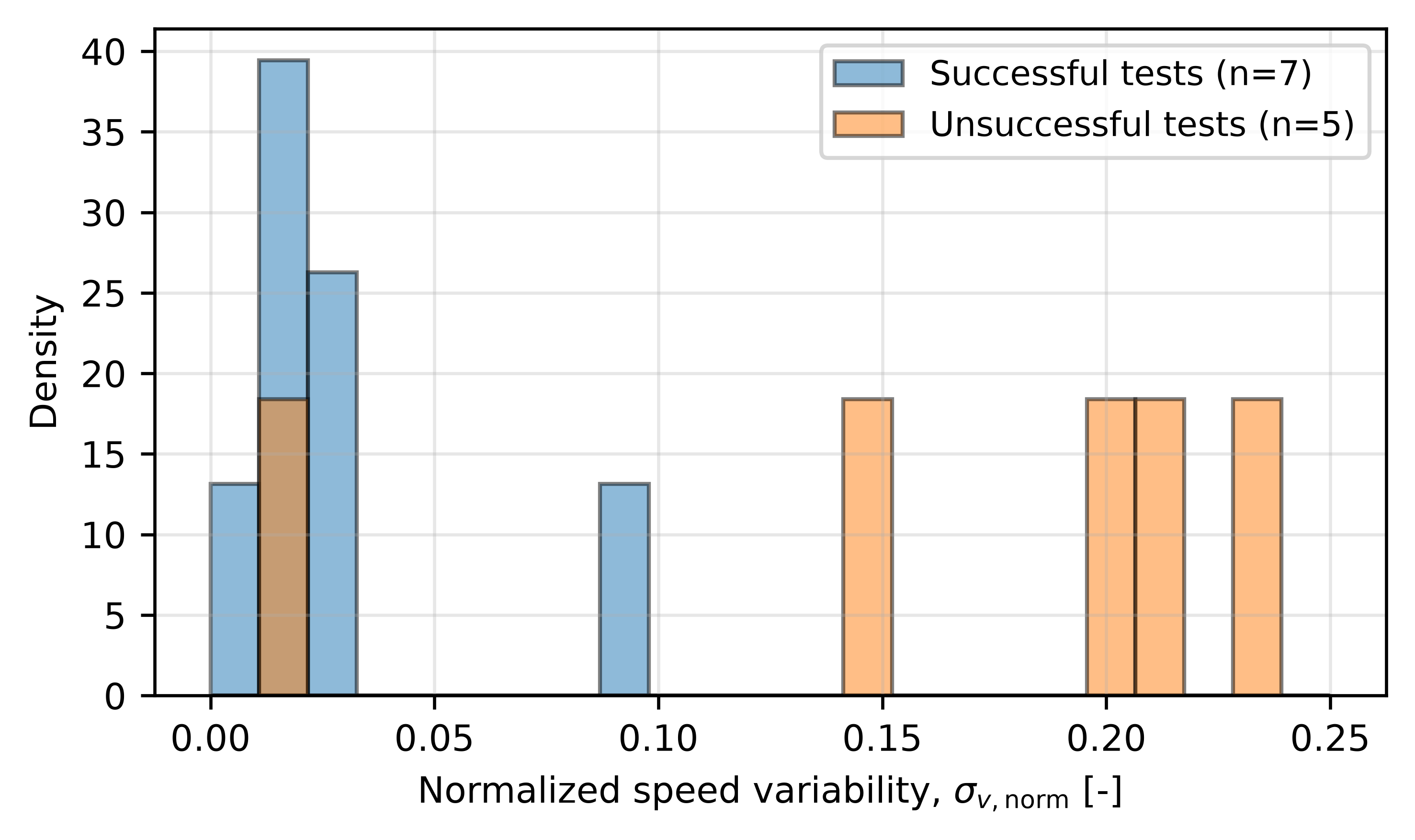}
    \caption{Empirical distribution of Normalized 
    Speed
    Variability for successful ($n=7$) and unsuccessful ($n=5$) AEBS tests.}
    \label{fig:ECDF}
\end{figure}

In a few cases, the AEBS failed to intervene despite the test run being within UN-R152 limits.
To investigate potential factors associated with unsuccessful AEBS interventions, the normalized standard deviation of VUT speed was compared between successful and unsuccessful test runs. As shown in \cref{fig:ECDF}, the two groups exhibit markedly different distributions of speed variability during the approach to the GVT. Successful tests showed a substantially more stable approach, with a mean normalized speed standard deviation of 0.028 ($n=7$), compared with 0.163 for unsuccessful tests ($n=5$). Thus, the average normalized speed variability was approximately six times higher in the unsuccessful cases.

The difference between the two groups was assessed using Welch's $t$-test. The test indicated a statistically significant difference in normalized speed variability ($t=-3.44$, $p=0.021$), with the unsuccessful tests exhibiting substantially greater variability. The estimated mean difference between the successful and unsuccessful groups was $-0.135$, with a 95\% Wilson confidence interval of approximately $[-0.236,-0.034]$. These results suggest that increased longitudinal speed variability during the approach is associated with unsuccessful AEBS intervention.
This finding should nevertheless be interpreted with caution given the limited number of observations. In particular, the analysis identifies an association rather than a causal relationship between speed variability and AEBS outcome. 


\section{Summary and Conclusions}\label{sec:conclusion}

This study experimentally characterized the longitudinal behavior of commercial AEBS and ACC in controlled rear-end scenarios.
The results showed clear differences in their intervention strategies: ACC exhibited a more progressive response, with mild braking initiated over a broader range of time-to-collision (TTC) values, whereas AEBS exhibited a predominantly reactive response, with pronounced braking concentrated at shorter TTC values.

In nominal tests, ACC achieved almost complete stopping success across the tested speed range, with some degradation above 80~km/h. AEBS achieved complete stopping success below 40~km/h, with less consistent outcomes at higher speeds.

Beyond conventional binary pass/fail assessment, the AEBS analysis found no statistically significant association between initial approach speed and unsuccessful intervention.  Given that tests were discontinued when the ADAS system became unstable due to safety of the car and drivers, which in turn resulted in a limited sample, the effect of speed cannot be ruled out under extended testing conditions.

In contrast, normalized speed variability differed significantly between successful and unsuccessful AEBS tests. This difference suggests that features of the approach speed profile, effects probably interpreted as driver interventions, may provide information on AEBS performance that is not captured by the initial speed alone, at least within the regulatory speed domain mandating AEBS intervention. Obviously, when the speed was fully controlled by the ACC system, such an anomaly was not observed, indicating safety benefits for ACC/L2 systems. 

The findings support complementing conventional speed-based assessment with trajectory- and test-level statistical analysis. The counterfactual injury-risk estimates further illustrate the potential safety implications of unsuccessful interventions, although they should not be interpreted as observed or population-level injury risks. These estimates assume a collision at the initial approach speed without further mitigation. 

Nevertheless, the limited number of tests and unsuccessful interventions makes these results exploratory. Furthermore, aggregation across two vehicles enables a function-level comparison but prevents vehicle-specific effects from being isolated. Future work will extend the dataset and investigate the joint influence of variables such as approach speed, lateral positioning, and speed variability.

\section*{ACKNOWLEDGMENTS}
The authors would like to thank RDW for their support, in particular, Nick Koetsveld and Gijsbert de Kam.
The Joint Research Centre for the European Commission supported this work. The opinions expressed are those of the authors and should not be considered to represent an official opinion of the European Commission.
During the preparation of this work, the authors used OpenAI GPT-5.6 Luna to improve the grammar. The authors reviewed the content as needed and take full responsibility.

\bibliographystyle{IEEEtran}
\bibliography{root} 
	
\end{document}